An Investigation into Active Control for
Accessible Orbital Flight

Timothy Cai

CSFJ 2022

University of British Columbia, BC, Canada

**Abstract**

Recently, a practical and publicly accessible satellite standard called the SmallSat has amplified public involvement in orbital research. This allows for flexible and efficient deployments of impactful low-earth-orbit experiments that would otherwise never be flown. However, the launch industry responsible for flying these experiments is not flexible nor efficient. This project aims to make orbital technologies accessible at the miniature scale, specifically thrust-vector-control, through an iterative engineering process simplifying and miniaturizing technologies from launch vehicles such as the Space Shuttle and Falcon 9. An Arduino-based custom flight computer was developed alongside state machine control software and active-control hardware, all designed to scale. Together, these three major components emulate the methods used in the aerospace industry. Initial test flights and recent ground test data have indicated stable control with a maximum of 7° and 2.62° of deviation from the intended flight path respectively, an acceptable stability range when compared to similar finned flights. Results show that scalable thrust vectoring is possible at a small scale, giving adaptability and control applicable to both small and large test vehicles. With accessible orbital flight, countless experiments can be completed concurrently, allowing for faster amateur rocket development and opening another path to space.

# 1 Introduction

## 1.1 Background

Placing satellites into orbit has always been restricted to private companies and those with vast resources. This monopoly is a bottleneck for scientific discovery and exploration; experimentation is limited to the objectives of those in control of the launch vehicles and satellites that allow for these activities. However, in the past decade, public involvement in space has exploded due to the introduction of a new satellite standard called the "SmallSat," a practical and publicly accessible form factor that allows anyone to design and build an orbital experiment (Crook, 2009). However, SmallSat launch opportunities have not increased enough to meet demand, as the number of launch providers capable of delivery to orbit has not grown accordingly (Hader, 2021). Even for companies with higher launch cadences such as SpaceX, the cost of the most economical rideshare possible is in the range of $150,000 (Spaceflight, 2020) to $1,000,000 (SpaceX, 2019), prohibitive to most hobbyists and universities. Commercial vehicles also do not have the capacity to deliver to specific inclinations and altitudes, restricting rideshare orbits (Clark, 2020). Dedicated SmallSat launchers are more flexible, but also scarcer and far more expensive (Sesnic, 2021). As a result of the current launch climate, community-driven scientific exploration outside of Low Earth Orbit (LEO) is extremely difficult.

One of the reasons why the public does not have their own way to access space is the lack of access to a particular technology known as active trajectory control. It is the largest barrier of entry for higher altitude flights, as small deviations in the flight path can cause a rocket to fail its orbital insertion. Without active control, a rocket is flying blind, and a great risk is involved in flying a payload. One approach to active trajectory control is thrust vector control (TVC). Exclusive to cost-prohibitive, large-scale orbital launch vehicles, TVC is the automatic movement and redirection of a rocket's thrust and allows for precise trajectories and reliable stability. It is the indispensable element that enables consistent, successful flights or even vertical landing. No active control is used with economical nano-scale orbital rockets such as the JAXA S-Series (JAXA, 2018), causing inaccurate orbit placements and occasional failures due to launch and stage-separation variation.

Miniaturized active control is still an emerging technology. If TVC is made accessible to the amateurs already building rockets across the world, public access to LEO would be paved.

**1.2    Purpose**

The goal of this project is to demonstrate a combination of advanced active control in economical and flexible small-scale rockets to meet growing demand for access to space. A working prototype was developed with accessible parts and manufacturing techniques. The goal is to have active stability in flight.

**1.3    Technical Objectives**

A successful flight is defined as a nominal trajectory with no more than five degrees of deviation from the intended flight path at any time; this is a realistic goal for the noisy and inaccurate commercial hardware being used (UBC-Rocket & Cai, T., Personal Communication, 2021). The rocket will be flown on an E12 Estes solid rocket motor as recommended by UBC Rocket (2021). TVC gimbal must be able to reach an angle of ±10° on any axis, as per NASA RS-25 specifications (Dumoulin & NASA, 2009), as well as being able to operate nominally during the motor's peak thrust (15N). The gimbal design itself must be created with scalability in mind, with as much similarity to industrial design as possible, for future use in larger, higher power rockets (HPR). At minimum, it must function properly at the model scale.

| Objective | Application | Goal | Importance |
|---|---|---|---|
| Stability | The goal of stability in TVC is to rival the stability of finned rockets, as well as provide control authority in space. | ±5° in flight, in the pitch and yaw axes. | High |
| TVC Gimbal | The better the gimbal is built, the greater control authority the control system has. This will allow the rocket to correct extreme situations. | ±10° in the pitch and yaw axes. At minimum, it must successfully redirect 15N of thrust. | High |
| Recovery | Any model rocket must have a recovery system in place. | Timely parachute ejection at specified altitude. | Medium |
| Flight Computer Board | The firmware provides all of the instructions to keep the rocket upright. It is critical to a stable flight and must run at a sufficient speed for flight. | Properly tracks orientation and provides fast and accurate responses. 30hz at minimum, and ideally at ~100hz. | Medium |
| Launchpad | The launchpad holds the rocket upright in preflight procedures and redirects the heat of launch. Additionally, it also lights the engine for liftoff. | Stabilizes the rocket before flight and deflects motor flames from the test site. Successfully controls the launch clamps and ignites the engine. | Low |

**Figure 1**
*A table of design requirements*

## 2 Methods and Materials

## 2.1 Design Process

To miniaturize large scale industrial concepts, commercial alternatives or custom designs are used according to the criteria described in 1.3. To meet the objectives effectively, iterative development (NASA, 2019) is employed; multiple revisions of hardware and software designs were created, tested, analyzed, and then improved upon, in accordance with part testing, simulation, and flight data. Throughout the course of this project, two major revisions were developed, constructed, and tested. These two revisions will be referred to as Revision 1 and Revision 2.

### 2.2.1 Active-Control Hardware

The central piece of hardware is the thrust-vector control gimbal mount. The design takes inspiration from the Aerojet Rocketdyne RS-25 and SpaceX Raptor. The final revision is a gimbal with a compliant joint printed with a flexible filament (thermoplastic polyurethane) and hobby servo motors in lieu of linear actuators stacked on top (Figure 2). The force of the servos is transmitted to the engine mount through 1-millimeter steel pushrods. The final version reaches a 15° range of motion in both the yaw and pitch axis.

Besides operational ability, the focus of the gimbal is scalability. The vertically oriented design of the gimbal allows larger designs to scale axially, rather than radially, optimal for high-powered rocket airframes that contain large amounts of vertical space. As the design matures and is developed for larger engines, the layout allows for simple replacement for higher strength servos or linear actuators.

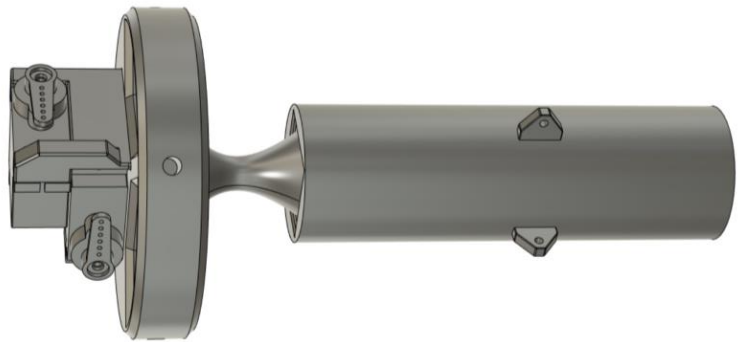

**Figure 2**
*The thrust vectoring gimbal*

### 2.2.2 Recovery Hardware

The ability to land the rocket at a safe speed is a critical hardware requirement. Both Canadian and National Rocketry Associations (CAR & NAR) require an appropriately sized parachute and method of parachute ejection to be flown on model rockets, and no flight in this project was without one. A dual-spring powered piston ejection system was utilized, with a peak ejection force of 50N. The speed of the piston during ejection was measured to be in the range of 17 meters per second, able to reliably release parachutes.

### 2.3.1 Electronics

The avionics on the rocket consist of a custom designed flight computer with servo outputs and flight sensors.

### 2.3.2 Flight Computer Version 1

The current design is a two-layer PCB that makes the connections between several commercial breakout boards, servo outputs and pyrotechnic channels using through-hole mounts, which makes for relatively simple assembly and component selection (Figure 3). It measures 100x66 millimetres in height and width. The design files were sent to a commercial PCB manufacturer and assembled in-house. Once assembled and tested, the computer was confirmed to be able to detect acceleration, control servos, and indicate stage changes with its LED and buzzer. This was the main flight computer for the first test campaign and flown on the Revision 1 and 2 test flights.

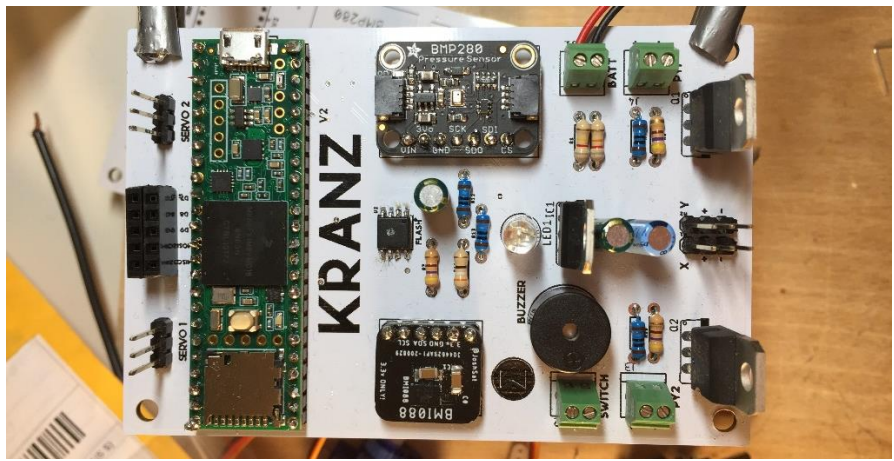

**Figure 3**
*The Kranz flight computer*

**2.3.3 Flight Computer Version 2**

The next-generation flight computer is a surface-mount (SMD) based board with over 120 components. The SMD board is four times smaller than FCV2, measuring 50x30 millimetres, and contains a far more robust sensor suite, with multiple accelerometers and GPS. This version of the flight computer is built to verify the project's first advanced SMD design under high power conditions, alongside a reaction control wheel and its accompanying control algorithms. A launch on an ESA Dual CanSat Launch Vehicle to an altitude of one kilometre verified its performance under load.

**2.4.1 Flight Control Firmware**

The flight control firmware is the bridge between the mechanical and electrical system. The design for the code was laid out according to Dumoulin & NASA (2009) and Tarazevits (2015). The most up-to-date codebase is now a fully assembled project in Visual Studio Code, the details of which are covered in later sections.

**2.4.2 Orientation and Coordinate Systems**

The raw angular velocity input was read from an Inertial Measurement Unit (IMU) using an Arduino library. To prevent inaccurate data from gyroscope drift while waiting on the pad, a gyroscope bias algorithm is implemented to increase accuracy of the angular velocity inputs. In order to transform the local coordinate system of the IMU to the global coordinates used by the control loop, a complex number system known as quaternions are used (Madgwick, 2010).

The advantages of quaternions include their low computation load and immunity to gimbal lock. They were utilized on the Space Shuttle as the primary orientation system, with a long history of usage in spaceflight (Schletz, 1982). Once converted to a Tait-Bryan (yaw, pitch, roll) orientation form, it was fed into the active control algorithm.

### 2.4.3 PID Control Algorithm

The Proportional, Integral, Derivative control system (PID) was chosen for many reasons. Its relative simplicity (in comparison with other control schemes such as full state feedback) and computational load make it an excellent choice for lower-powered flight computers, and its wide range of use make it very accessible with large amounts of documentation.

The PID gains were originally determined in a simulated environment for Revision 1 that accounted for wind, servo offsets, motor thrust curves, and uneven launches. Eventually, Revision 2 verified simulated gains through physical ground testing. Both simulated and physical tuning processes employed the Zeigler-Nichols method, a heuristic method of manually tuning PIDs (Bennett, 2022).

### 2.4.4 Data Acquisition

Data collection is critical to the project, and every sensor and setting is recorded for each flight and test. Everything from the raw and processed gyroscope data, system state, and flight settings like PID gains and gyro biases are recorded in a separate file. Data is logged at 1 hertz on the pad, and at 60 Hz in powered flight, and saved onto an SD card post-flight.

### 2.4.5 Finite State Machine

The firmware was structured as a finite state machine, similar to the one used on the Falcon 9 (Tarazevits, 2015) and Space Launch System (Harris et al., 2013). This is a common control scheme used in many industry rockets and contains major flight stages. The active flight control is activated only in "Powered Ascent". The full state machine can be seen in Figure 4.

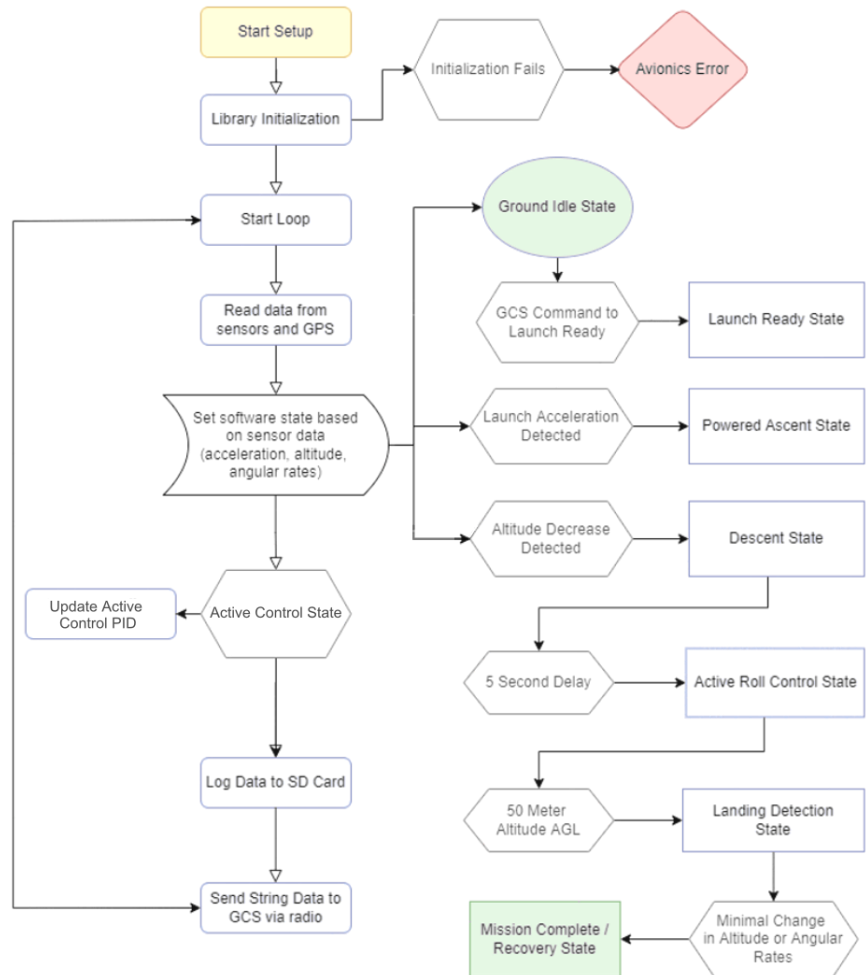

**Figure 4**

*A flowchart of different software states of the test vehicle*

### 2.5.1 Primary Testing Processes

Test flights are integral to the verification of the control system and launches had to follow Canadian Association of Rocketry rules, as well as some additional recommendations from NAR. Each test provided a wealth of data, especially in accordance with simulation data and camera video, and each data point was measured and compared to the design criteria. After analyzing the recorded data, changes can be made to the design, and the rocket can then be rebuilt and flown again. The iterative design process is summarized in Figure 5.

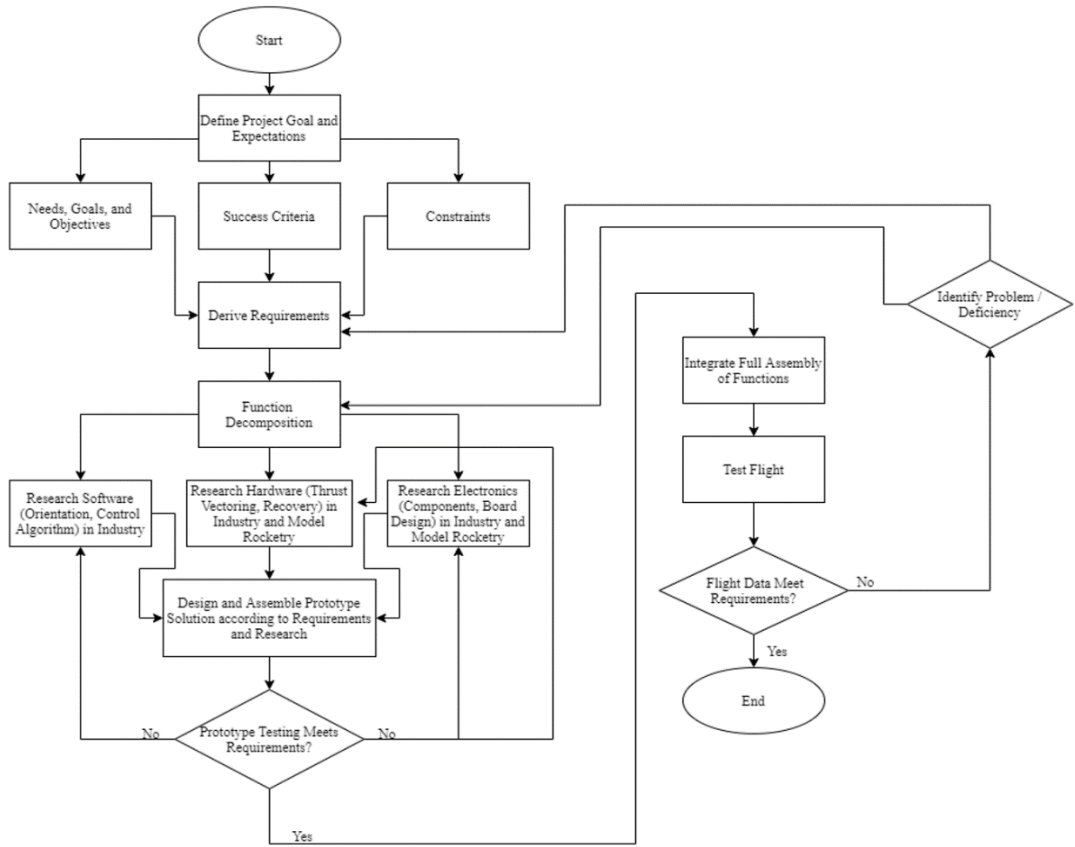

**Figure 5**
*A detailed flowchart of the iterative design process*

After the four flights of Revision 1, iteration inefficiencies associated with the complex failure modes in flight were recognized. Multiple aspects of the rocket, from the servos to the PBC power delivery to the software states, could fail in the air and were difficult to diagnose. Thus, a new process was implemented to better isolate the rocket's systems (Figure 6). A high-powered brushless DC motor is used to produce one-third thrust on the ground and verify the rocket's control systems and component integration without risk. This is currently the only method used to evaluate Revision 2, which is progressing to a test flight.

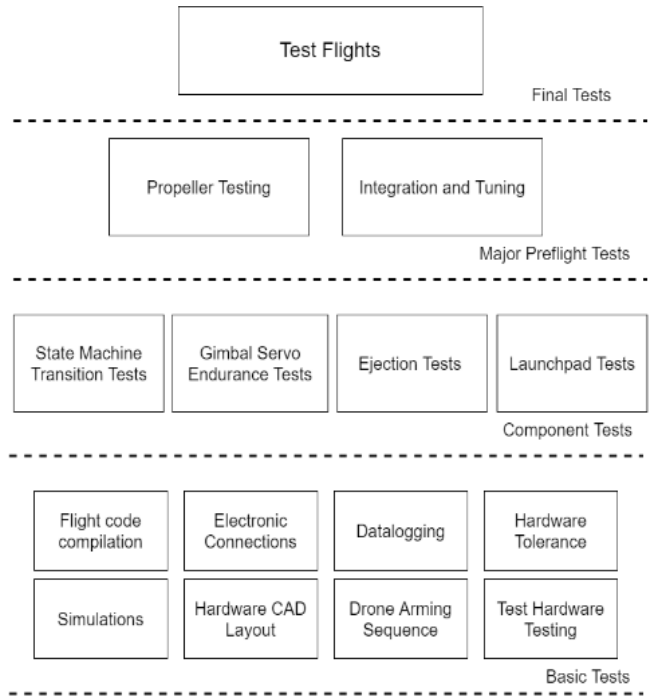

**Figure 6**

*A hierarchy of verification tests for the test vehicle*

## 3    Results

In the most successful flight of the Revision 1 test campaign, flight orientation was close to nominal with a maximum of 7 degrees of deviation and a response time of 0.86 seconds from course deflection to correction. Once the flight data was compared to simulation data, it was found that the motor underperformed due to humidity and the additional weight of an onboard camera; however, the rocket and its control systems were able to achieve nearly vertical flight.

In the most successful ground test of Revision 2, a maximum deviation of 2.62 degrees was recorded, with a response time of 0.48s for the rocket to right itself from a minor redirection (Figure 7).

## 4    Discussion

Control in flight has been measured to be insufficient, yet the systems and methods indeed do have a measurable and positive effect on launches. Several flights were very close to meeting requirements, and ground testing was successful. In reference to Figure 1 (flight requirements), every flight criterion was met except stability. Some flights were 90.0° over the stability goal, but the most successful were closer to 2.0° over the stability target. In ground testing, stability has been consistently 3.0° underneath the threshold. The issue now lies in the reliability of the control system. Faults in the tuning process, oversights in the code and state machine, as well as loose tolerances in the thrust vectoring mount all contribute to the lack of precise control during flight.

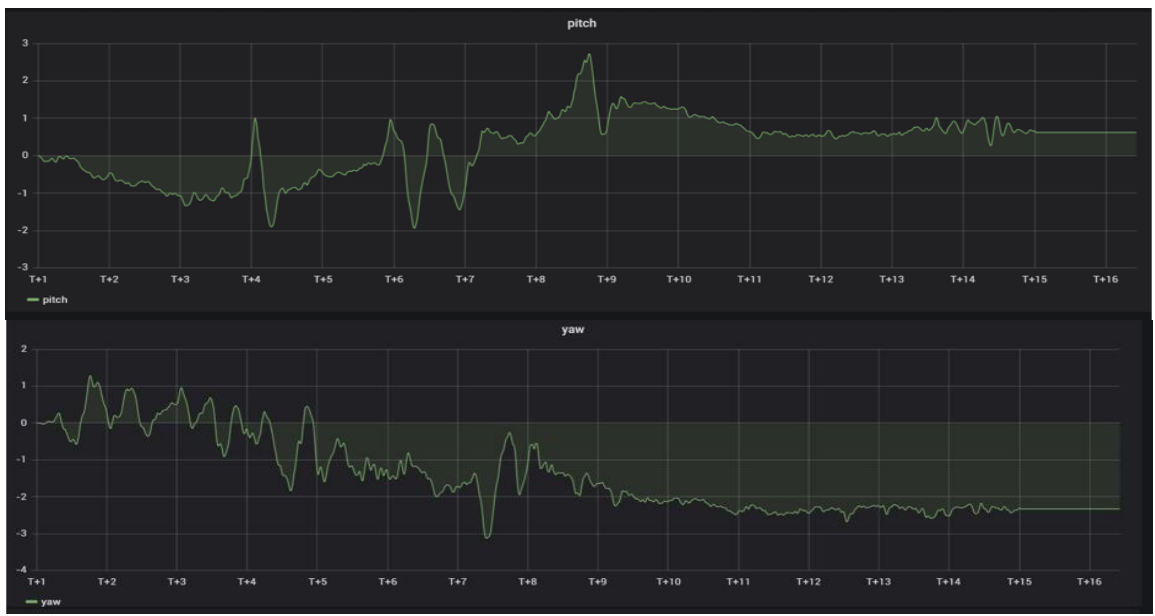

**Figure 7**
*Data from a Revision 2 ground test*
*x axis is time on both graphs.*
*Upper graph is pitch; lower graph 2 is yaw.*

## 5   Conclusion

As with any other test campaign, there were some problems in the hardware and software, but the fundamental operation of the rocket is valid. The state machine worked on commercial hardware, the quaternion orientation accounted for yaw, pitch, and roll, and the control system successfully kept the rocket mostly upright with the gimbal. Scalable active control on the miniature scale is viable. This allows for more applicable TVC test vehicles in the amateur space, and eventually, greater access to orbit for all.

Citations